# Evaluating Protein-protein Interaction Predictors with a Novel 3-Dimensional Metric

Haohan Wang[1], Madhavi K. Ganapathiraju, PhD[2]

[1]Language Technologies Institute, School of Computer Science, Carnegie Mellon University, Pittsburgh, PA;

[2]Department of Biomedical Informatics, University of Pittsburgh, Pittsburgh, PA

**Abstract**

*In order for the predicted interactions to be directly adopted by biologists, the machine learning predictions have to be of high precision, regardless of recall. This aspect cannot be evaluated or numerically represented well by traditional metrics like accuracy, ROC, or precision-recall curve. In this work, we start from the alignment in sensitivity of ROC and recall of precision-recall curve, and propose an evaluation metric focusing on the ability of a model to be adopted by biologists. This metric evaluates the ability of a machine learning algorithm to predict only new interactions, meanwhile, it eliminates the influence of test dataset. In the experiment of evaluating different classifiers with a same data set and evaluating the same predictor with different datasets, our new metric fulfills the evaluation task of our interest while two widely recognized metrics, ROC and precision-recall curve fail the tasks for different reasons.* [1]

## 1. Introduction

Identifying protein-protein interaction pairs is a fundamental step for investigating protein functions and understanding the inherent biological processes of human. While there is a great need for more laboratory experiments to be carried out in discovering new interactions, researchers have to face the many constraints, limitations of equipment or subjects slow down the discovering processes. Meanwhile, machine learning based solutions are not constrained with physical environment. As long as data is presented, a machine learning model can be trained to predict new interactions.

Because of this advantage of computational models, a great number of machine learning methods are introduced to predict protein interactions to accelerate laboratory research. Most of them are found to be promising with one or several traditional evaluation metrics, like accuracy, ROC [3-4], or precision recall curve [5]. However, because of the fact that each of the verification of a single predicted interaction in laboratory is very expensive, researchers expect the machine learning model can predict positive interactions with as less noises of non-interactions as possible; while researchers don't usually expect the prediction result of non-interactions as pure as possible since no one will verify predicted non-interactions. When a machine learning model is applied to predicting interactions, researchers expect the model to be focused on the ability of predicting new interaction pairs, rather than on the ability of excluding non-interaction pairs. In other words, researchers can benefit from high precision machine learning model, of which the verification requires little effort in laboratory experiments; while recall of a machine learning model is not very essential. Because of this real world usage requirement for PPI predictors, many machine learning models that are shown promising with some evaluation metrics may not function as well as claimed when applied by laboratory researchers.

---

[1] This article is an extended version of a poster presented in AMIA TBI 2015

There are many predictors proposed considering this real world usage. Most of these machine learning models are evaluated with precision recall curve, with a long passage explaining the behavior changes along precision recall curve and showing these predictors can predict what the researchers expect. While this evaluation method can fulfill the task, there is still an obvious drawback: there is not a numeric value that can be used to compare across all these models. Figures are very illustrative, but not very comparable. Another drawback of this evaluation metric is that the result partially contains the information of test data. The ability of predicting positive interactions is highly affected with the percentage of positive interactions in the test data set. A predictor evaluated a precision of 0.99 at recall=0.2 may not be better than a predictor showed a precision of 0.8 at recall=0.2 simply because there is no standard criteria constraining the distribution of test data.

In this work, we proposed a new evaluation metric for intearctomes predicting machine learning models. Our new evaluation metric is aimed to solve the two problems we mentioned above: 1) *High precision*. We mainly focus on the ability of minimizing the number of wrong predictions when predicting new interactions, and neglect the ability of minimizing false negative. 2) *Test data independent*. A machine learning model should be evaluated consistently with no relevance of the distribution of different classes of test data, so we tried to exclude the impact of test data.

Besides these two main goals, we also consider several general criteria for an evaluation metric mentioned by Anagnostopous et al [1]: 1) *Simple intuition*. An evaluation metric must be simply understood for being widely used. 2) *Simple computation*. The result of a metric has to be tractable with reasonable power of computation for the scale of the problem set interested, for being widely used. 3) *Simple representation*. The evaluated result should be represented by a simple number, thus allowing comparison across different models in different work.

The metric proposed in this paper is built on these criteria. The following part of this paper is organized as follow: First, we will briefly talk about other evaluation metrics and why they cannot perform our evaluation goal. Then, we will derive our metric and introduce the intuition behind it. Next, we will compare our metric with two main metrics widely used, ROC and precision recall curve, under a simple probabilistic partial oracle predictor and toy data. Finally, a conclusion is drawn and future works are stated.

## 2. Different Evaluation Metrics and their Constraints in Evaluating Real World Interactions Predictors

In this section, we will talk about different existing evaluation metrics and the reasons for their drawbacks. We will first focus on the three most widely recognized metrics: accuracy, ROC and precision-recall, then briefly talk about other metrics. Confusion matrix[2], as showed in Table 1, will be repeatedly used across these metrics.

**Table 1.** A Confusion Matrix. The columns show the gold standard labels defined by the data set and rows show the labels predicted by models.

|  | **Gold standard positive** | **Gold standard negative** |
|---|---|---|
| Predicted positive | True positive (tp) | False positive (fp) |
| Predicted negative | False negative (fn) | True negative (tn) |

Accuracy is one of the most used evaluation metric of predicting tasks. It can be simply explained as the percentage of correctly predicted labels, regardless of positive or negative. It is defined as $(tp+tn)/(tp+fp+fn+tn)$. The limitation is very straightforward. For a protein-protein interaction research, 99.9% of the data will be negative data (data of no interest), thus, simply predicting everything to be non-interactions will result in an accuracy of almost 1.0. Thus, accuracy cannot capture the real world adopted predicting ability of a model.

ROC (Receiver Operating Characteristic) curve is a plot characterized by sensitivity (defined as $tp/(tp+fn)$) and 1-specificity (defined as $1-tn/fp+tn$) of a model with a set of varied thresholds for prediction. It describes the performance of a binary classifier very well and it comes with a numeric score called AUC (area under ROC curve). However, in the most widely recognized interpretation of ROC applied to evaluating protein interaction model, it evaluates regardless of the costs of two types of errors [6], emphasizing both the abilities of predicting new interaction and excluding non-interaction equivalently. Thus it is not a good evaluation metric to select the models to be adopted by biologists.

Precision recall curve is a plot characterized by different set of precision (defined as $tp/tp+fp$) and recall (defined as $tp/tp+fn$) of the model evaluated with different thresholds selected. It mainly captures the precision and recall tradeoff of a model. Area under precision recall curve is not very widely, but this numeric value is still used in some work comparing ROC and precision recall curve [7-8]. Although it is designed to describe the ability of predicting new interactions with different ability of excluding non-interactions, future sections will show that precision-recall curve is highly affected by the distribution of test data set.

Cost curve [9-10] is a point/line duality of ROC curve. Different costs are set for type 1 error and type 2 error. Thus, it could differentiate the abilities of predicting new interactions and excluding non-interactions. However, the drawback of cost curves are very straighforwad. There are no illustrative figures and numeric values for comparison. H-measure [11] is a special case of cost curve, what basically shows the same ability and constraints of cost curve. [12]

AUK (Area Under Kappa) [13] is another metric that focuses on the Kappa statistics [14]. It is designed to capture the ability of predicting minority class, however, it is still test data dependepent.

### 3. New Evaluation Metric

In this section, we will introduce the evaluation metric that we proposed. Before that, we need to establish one assumption: the machine learning algorithm we evaluate has to predict at least one positive interaction, regardless of whether this predicted interaction is positive or negative in gold standard. This is a reasonable assumption, especially considering the real world usage of directing biological research: a machine learning model predicting everything to be non-interactions is certainly useless. However, this reasonable assumption plays an important role in explaining many phenomenon.

Our method is derived based on the two most famous criteria, ROC and precision recall curve. Instead of traditional precision, we only focus on the purity of predicted interactions, which means that we focus to evaluate the portion of non-interactions contained in the predicted interaction set. A simple metric is applied that, instead of working on maximizing traditional precision, which defined as *tp/tp+fp*, we work on maximizing the ratio of true interactions predicted and non-interactions predicted, i.e. *tp/fp*. It is simple to notice that this ratio can be re-written with precision as *precision*/(1-*precision*). However, there is an essential difference between this ratio and precision. Precision is lower bounded by the portion of positive data in test data. The lower bound is reached when we select the threshold to predict everything to be positive, but we cannot set a threshold to predict everything to be negative because of the weak assumption. However, the ratio of true positives and false positives fall into the closed interval between 0 and infinity. There is no inherent constraint of this ratio. Besides this advantage, this ratio highly stresses the importance of purity in positive class prediction. For the same number of true positive interactions predicted, 2 false positive prediction will result in only half of the score of a prediction with only 1 false positive. This shows a clear contrast with precision, which is designed to be linearly interpreted. This is the first dimension of our metric space.

One flaw of the first dimension is that the tp/fp ratio cannot differentiate the predictors predicting only a few true positive interactions and even fewer false positive interactions from the predictors predicting many false positive interactions and even more true positive interactions. Another dimension of our metric space is introduced to measure the percentage of false positive interactions predicted out of all the gold standard negative interactions. We use the fp/tn ratio to capture this information. However, since the predictors are aimed to minimize this fp/tn ratio, we use 1-*fp*/*tn* as the second dimension. Similar to first dimension, this ratio can be represented by existing evaluation criterion, specificity. The ratio can be represented by -1/*specificity*.

A third dimension is introduced to our evaluation metric as the links between first two dimensions. Although our goal is to maximize the purity of predicting positive interactions, we won't deny the fact that for the same purity, a classifier of a better purity in negative prediction is better. Thus, this third dimension of our metric space serve as more than a link between the first two dimensions, it shows the traditional recall of the predictor. However, in this dimension; we don't have to derive a ratio because there is no need to exaggerate the importance of purity in negative predictions. Only linear interpretation should be enough.

Despite the fact that we have a 3D representation of the evaluation metric, the score is calculated only within 2 2D planes. The evaluation score is a product of the area constrained with tp/fp ratio and recall with the area constrained by fp/tn ratio.

### 4. Evaluation on Simple Probabilistic Partial Oracle Predictor and Toy Data

In this section, we will first introduce the simple predictors we design, and then we will compare our new evaluation metric with the two most widely recognized metrics, ROC and precision-recall curve on these simple predictors with some data we designed. We will show the advantage of our evaluation metric with these simple examples.

First, let's define $P(\alpha, \beta)$ as our predictor, predicting a probability for each instance as the probability to be a positive instance. It predicts deterministically correct label for every positive data with probability $\alpha$, and generates a less confident prediction for a positive data with probability 1- $\alpha$. It also predicts deterministically correct label for every negative data with probability $\beta$ and generates a less confident prediction for negative data with probability 1- $\beta$. Thus, $0 \leq \alpha, \beta \leq 1$. A deterministic prediction is a prediction with a confidence score of 1 for positive data and 0 for negative data. A less confident prediction only makes a random decision with the probability with some preference of the true label. The predictor generates the prediction result with a uniform distribution within [0.25, 1] for positive data and with a uniform distribution within [0, 0.75] for negative data. A perfect $P(\alpha, \beta)$ will be $\alpha=1$ and $\beta=1$, an increasing $\alpha$ and a decreasing $\beta$ can both result in more instances predicted to be positive with increasing true positive instances and increasing false positive instances respectively. Similarly, a decreasing $\alpha$ and an increasing $\beta$ will result in less instances predicted to be positive, more to be negative.

One may argue about the unrealistic assumption of this predictor that allows the predictor to behave with knowledge of test labels. Here we only adopt this mechanism to simulate the changes of true positives, false positives, true negatives and false negatives, thus we can use this $P(\alpha, \beta)$ to represent real world machine learning algorithms. Comparisons are only performed with these predictors in this section.

For comparison of our evaluation metric with ROC and precision-recall curve, we first evaluate the same predictors with different distribution of test data set and then we evaluate different predictors with the same distribution of data set. Then, we mainly compare the representation power of the numeric scores with help of the figures. For each evaluation metric, we evaluate four times as following: a) Evaluate the performance of P(0.1, 0.1) with five different data sets with different portion of positive data, 5%, 10%, 15%, 20%, 25%. b) Evaluate different models of an increasing α. c) Evaluate different models of a decreasing β. d) Evaluate different models of an increasing α and a decreasing β, while the sum of α and β stays the same. Because that our test data is set to be imbalanced (1 out of ten is positive), even the sum of α and β stays the same, these models does not behave the same. Smaller α and bigger β indicates the model predicting more instances correctly while bigger α and smaller β is favored because the model predicts more positive instances, although with increased noise.

The evaluation result with precision recall curve is shown in Figure. 1. As we can see that different abilities of predicting positive interactions can be captured very well by Figure 1(b) and Figure 1(c). However, the main flaw of this evaluation is that the result is clearly related with the portion of test data, which will stop researchers from comparing models across publications. The detailed score of each set of environment is shown in Table. 2. It shows clearly that the score is affected by portion of test data set. (Evaluation set a)

**Table 2.** Area under precision recall curve for different set of evaluations.

|  | Experiment Set#a | Experiment Set#b | Experiment Set#c | Experiment Set#d |
|---|---|---|---|---|
| Score #1 | 0.496 | 0.531 | 0.616 | 0.589 |
| Score #2 | 0.557 | 0.557 | 0.606 | 0.606 |
| Score #3 | 0.614 | 0.586 | 0.595 | 0.624 |
| Score #4 | 0.656 | 0.611 | 0.584 | 0.637 |
| Score #5 | 0.695 | 0.640 | 0.577 | 0.655 |

The evaluation result of ROC curve shows the information we expect to offer to biology laboratory researchers in the first three figures. However, in the forth figure, there is no clear result across the comparison. The AUC score is showed in Table 3, where we can see that there is a different for the models in Figure 2(d). However, a predictor with higher β is evaluated as a better predictor, which corresponds to the goal of ROC: evaluating the general performance of a machine learning model, regardless the difference of labels and the difference of two types of errors. Here we only concern the ROC as the one generally understood and applied across many interactions prediction works.

**Table 3.** Area under ROC for different set of evaluations.

|          | Experiment Set#a | Experiment Set#b | Experiment Set#c | Experiment Set#d |
|----------|------------------|------------------|------------------|------------------|
| Score #1 | 0.819 | 0.811 | 0.889 | 0.883 |
| Score #2 | 0.820 | 0.820 | 0.880 | 0.880 |
| Score #3 | 0.819 | 0.829 | 0.870 | 0.877 |
| Score #4 | 0.819 | 0.840 | 0.860 | 0.874 |
| Score #5 | 0.820 | 0.850 | 0.849 | 0.874 |

The metric of our evaluation metric is showed in Figure 3. However, since we didn't design our metric to be visually interpretable, we won't spend time here talking about figures. The numeric score is showed in Table 4. Most of our result showed consistently with what we can get from ROC curve, however, in set d, we don't treat those two types of error indifferently. Our ratio focuses more on the ability to predicting positive interactions with while constrained with the hazards of false positive. Thus, for set d, we rank the model that predicts more positive interactions highly instead of rank the one with general good performance higher.

**Table 4.** Numeric score calculated with our evaluation metric.

|          | Experiment Set#a | Experiment Set#b | Experiment Set#c | Experiment Set#d |
|----------|------------------|------------------|------------------|------------------|
| Score #1 | 0.247 | 0.236 | 0.322 | 0.297 |
| Score #2 | 0.253 | 0.251 | 0.311 | 0.314 |
| Score #3 | 0.251 | 0.273 | 0.305 | 0.322 |
| Score #4 | 0.254 | 0.290 | 0.292 | 0.332 |
| Score #5 | 0.252 | 0.308 | 0.289 | 0.341 |

## 5. Conclusion

In this work, we start from the goal of machine learning models applied to accelerate laboratory biology research, propose an evaluation metric that can describe the ability of a predictor to be adopted by biology researchers. We stress on that a machine learning model should predict positive interactions with very high confidence, thus that efforts are saved in laboratory. We propose a new evaluation metric focused mainly on the true positive and false positive rate, thus releasing the effect of positive data portion, considering false positive/true negative rate and recall of a classifier, that can select the best positive interactions predictors. We evaluate our evaluation metric with some toy machine learning models and showed the experiment result is consistent with what we believe.

In the future, we will apply our evaluation metric to evaluate some real world protein predictors, and make a better comparison across these predictors for biology research use.

**Acknowledgment**

This work has been funded by the Biobehavioral Research Awards for Innovative New Scientists (BRAINS) grant R01MH094564 awarded to MKG by the National Institute of Mental Health of National Institutes of Health (NIMH/NIH) of USA.

**References**

1. Anagnostopoulos, C. Measuring classification performance: the hmeasure package. 2012.
2. Stehman, Stephen V. Selecting and interpreting measures of thematic classification accuracy. Remote sensing of Environment 62.1 1997: 77-89.
3. Hanley, James A., and Barbara J. McNeil. The meaning and use of the area under a receiver operating characteristic (ROC) curve. Radiology 143.1 1982: 29-36.
4. Fawcett, Tom. An introduction to ROC analysis. Pattern recognition letters27.8 2006: 861-874.
5. Powers, David Martin. Evaluation: from precision, recall and F-measure to ROC, informedness, markedness and correlation. 2011.
6. Hand, David J. Evaluating diagnostic tests: the area under the ROC curve and the balance of errors. Statistics in Medicine 29.14 (2010): 1502-1510.
7. Davis, Jesse, and Mark Goadrich. The relationship between Precision-Recall and ROC curves. Proceedings of the 23rd international conference on Machine learning. ACM, 2006.
8. Boyd, Kendrick, Kevin H. Eng, and C. David Page. Area Under the Precision-Recall Curve: Point Estimates and Confidence Intervals. Machine Learning and Knowledge Discovery in Databases. Springer Berlin Heidelberg, 2013. 451-466.
9. Drummond, Chris, and Robert C. Holte. Cost curves: An improved method for visualizing classifier performance. (2006).
10. Drummond, Chris, and Robert C. Holte. What ROC Curves Can't Do (and Cost Curves Can). ROCAI. 2004.
11. Hand, David J. Measuring classifier performance: a coherent alternative to the area under the ROC curve. Machine learning 77.1 (2009): 103-123.
12. Ferri, Cèsar, José Hernández-Orallo, and Peter A. Flach. A coherent interpretation of AUC as a measure of aggregated classification performance. Pro-ceedings of the 28th International Conference on Machine Learning (ICML-11). 2011.
13. Kaymak, Uzay, Arie Ben-David, and Rob Potharst. The AUK: A simple alternative to the AUC. Engineering Applications of Artificial Intelligence 25.5 2012: 1082-1089.
14. Banerjee, Mousumi, et al. Beyond kappa: A review of interrater agreement measures. Canadian Journal of Statistics 27.1 (1999): 3-23.

**APPENDIX A: High Resolution Version of Figures.**

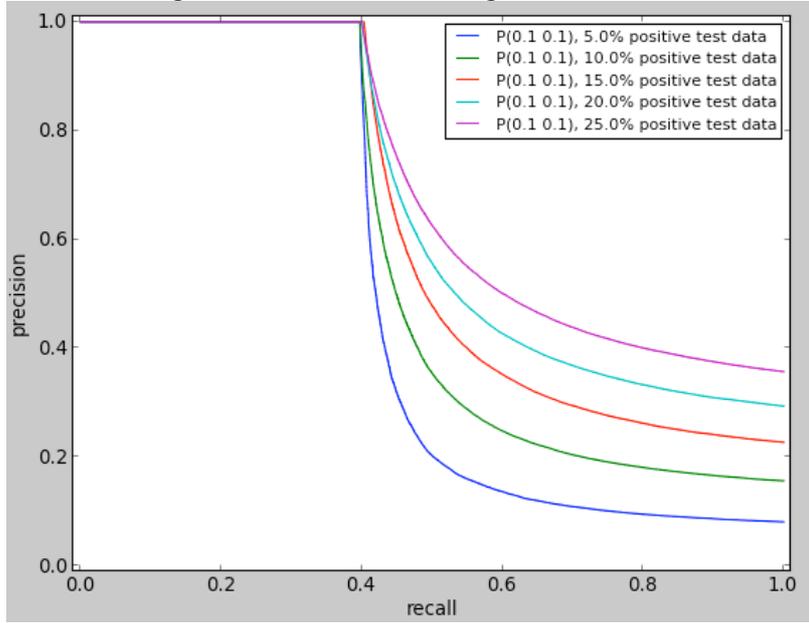

Figure 1 (a)

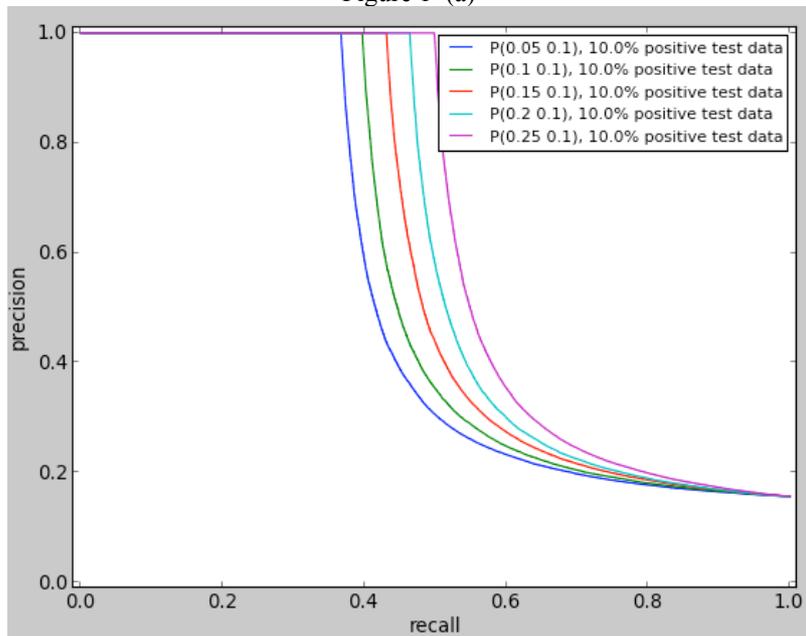

Figure 1(b)

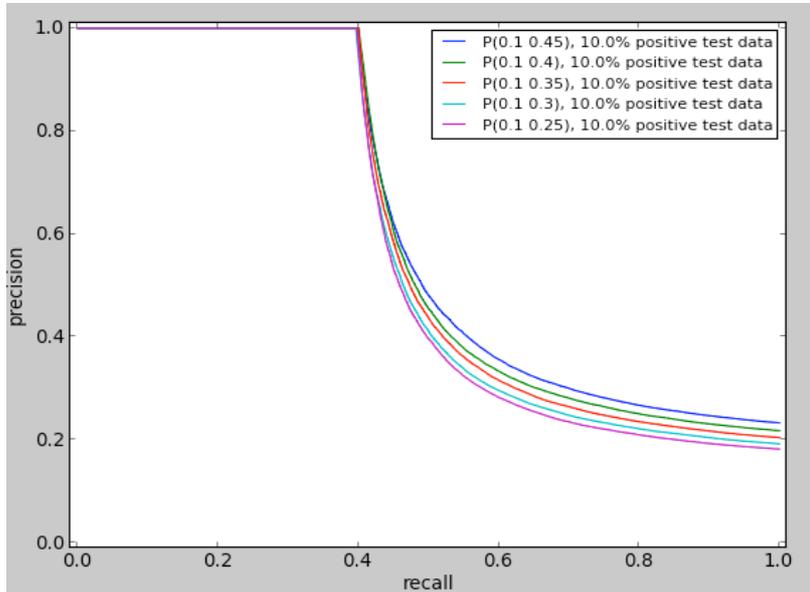

Figure 1(c)

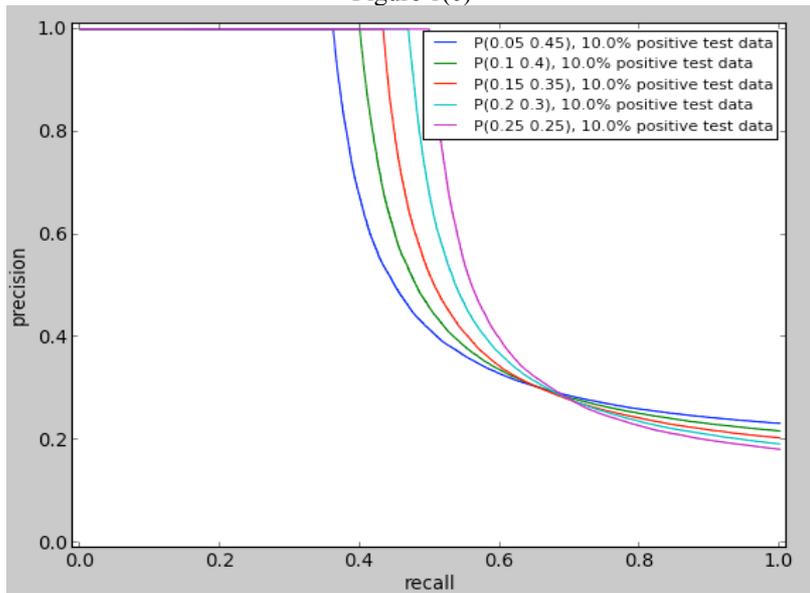

Figure 1(d)

**Figure 1**. Models evaluated with precision recall curve.

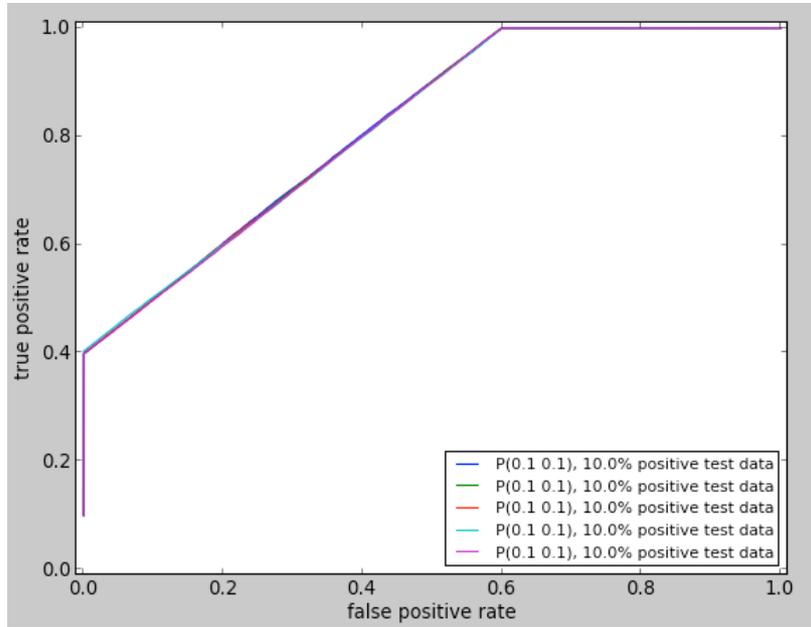
Figure 2(a)

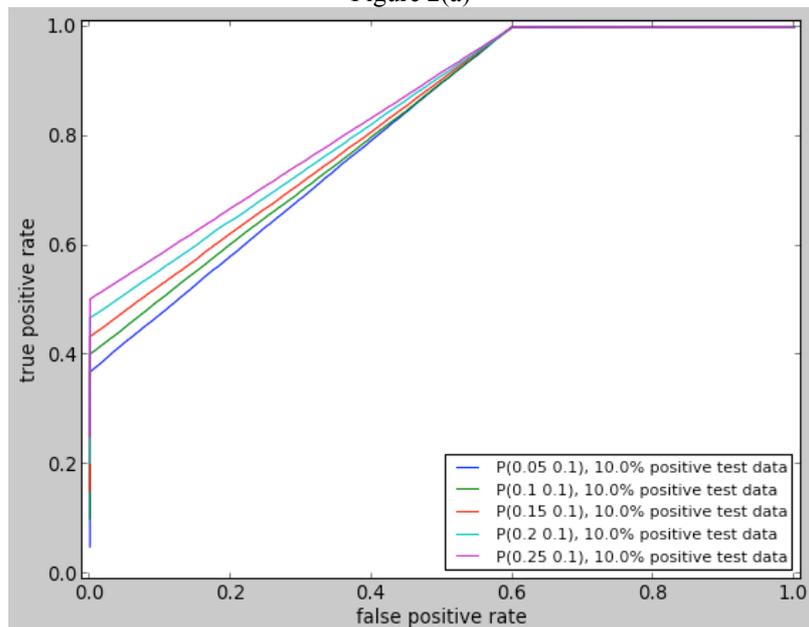
Figure 2(b)

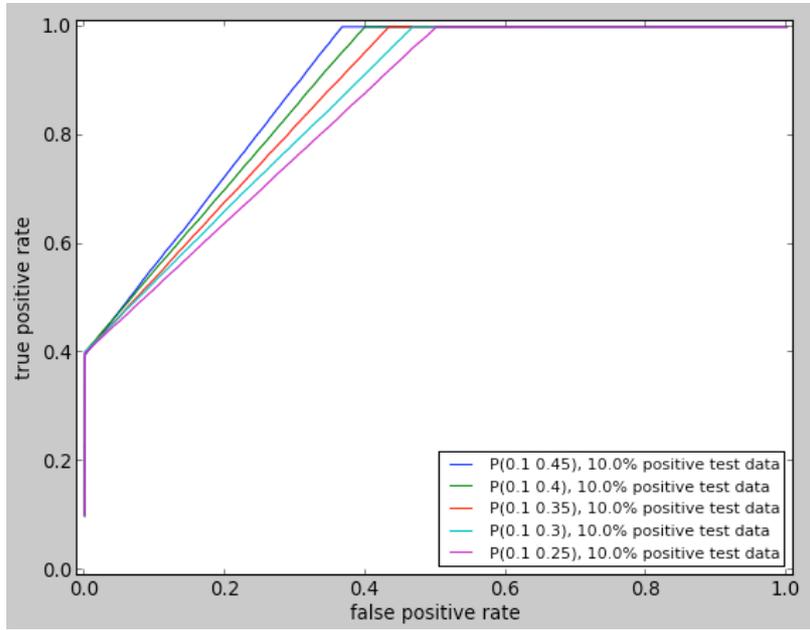
Figure 2(c)

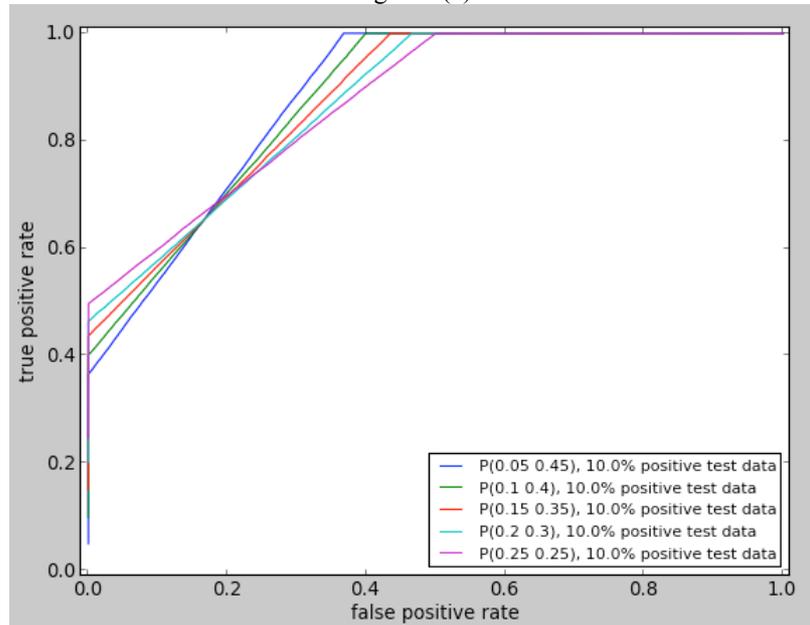
Figure 2(d)

**Figure 2**. Models evaluated with ROC

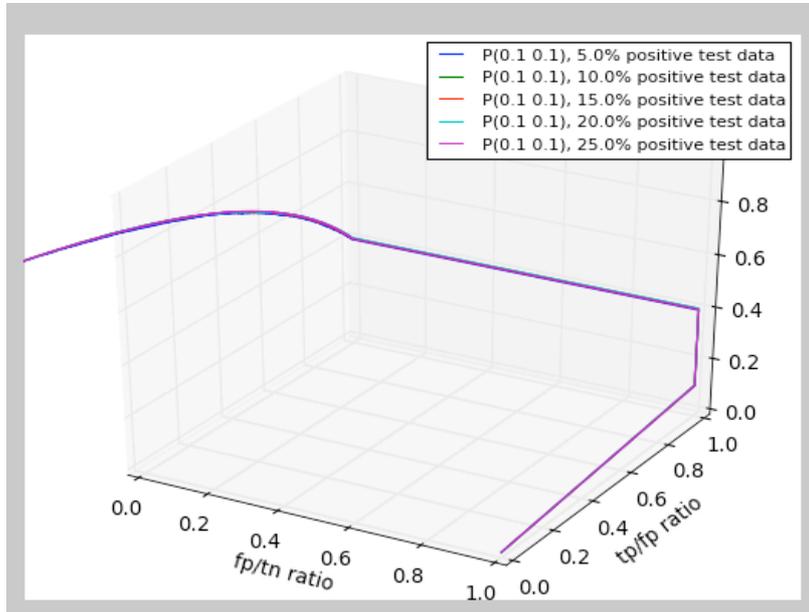

Figure 3(a)

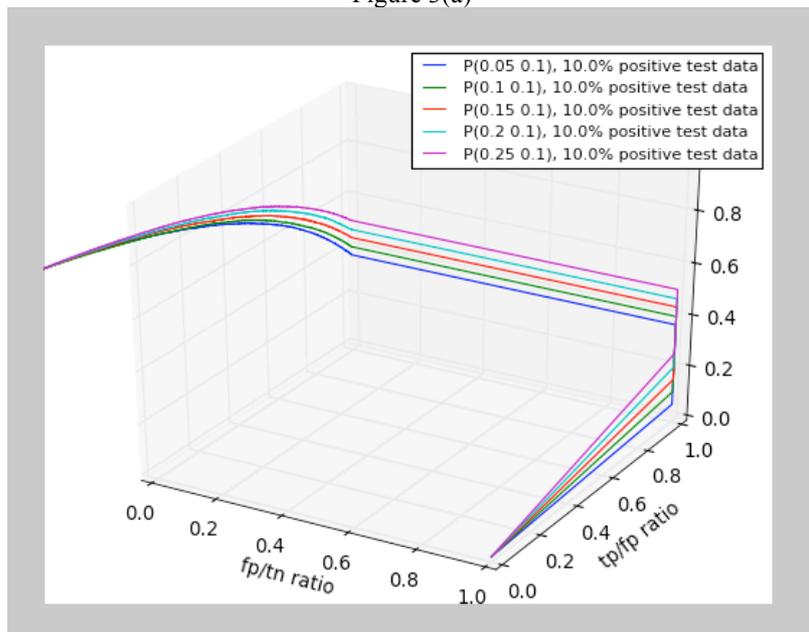

Figure 3(b)

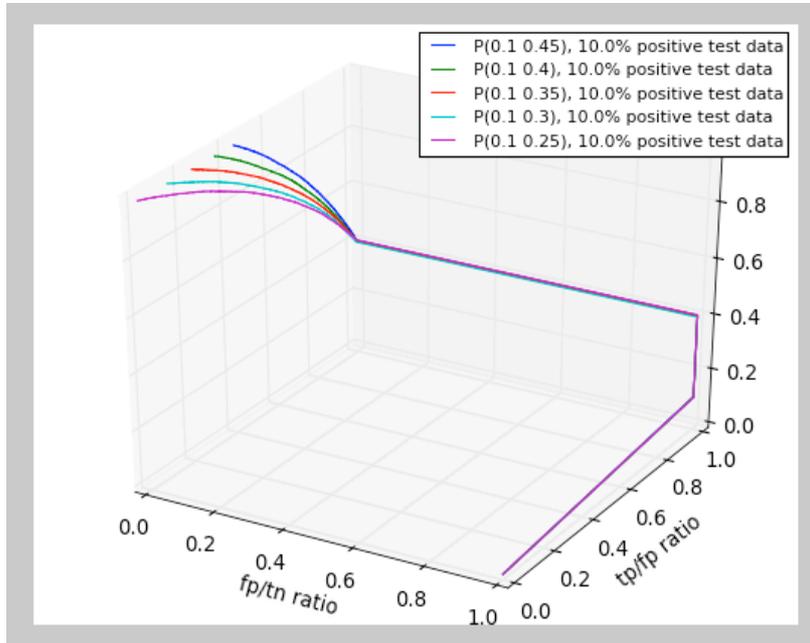

Figure 3(c)

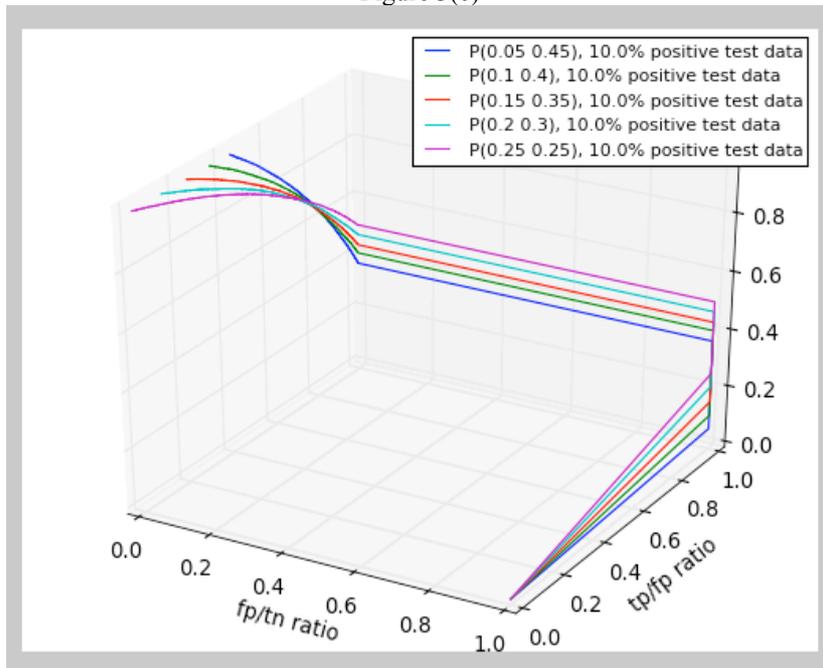

Figure 3(d)

**Figure 3**. Models evaluated with our new metric